\documentclass[sigconf]{acmart}

\AtBeginDocument{%
  \providecommand\BibTeX{{%
    \normalfont B\kern-0.5em{\scshape i\kern-0.25em b}\kern-0.8em\TeX}}}

\copyrightyear{2022} 
\acmYear{2022} 
\setcopyright{acmcopyright}\acmConference[SIGIR '22]{Proceedings of the 45th International ACM SIGIR Conference on Research and Development in Information Retrieval}{July 11--15, 2022}{Madrid, Spain}
\acmBooktitle{Proceedings of the 45th International ACM SIGIR Conference on Research and Development in Information Retrieval (SIGIR '22), July 11--15, 2022, Madrid, Spain}
\acmPrice{15.00}
\acmDOI{10.1145/3477495.3531697}
\acmISBN{978-1-4503-8732-3/22/07}

\begin{document}
\fancyhead{}
\title{SpaceQA: Answering Questions about the Design of Space Missions and Space Craft Concepts}

\author{Andres Garcia-Silva}
\email{agarcia@expert.ai}
\orcid{0000-0002-5664-488X}
\affiliation{%
  \institution{expert.ai}
  \streetaddress{Poeta Joan Maragall 3}
  \city{Madrid}
  \country{Spain}
  \postcode{28036}
}

\author{Cristian Berrio}
\email{cberrio@expert.ai}
\affiliation{%
  \institution{expert.ai}
   \streetaddress{Poeta Joan Maragall 3}
  \city{Madrid}
  \country{Spain}
  \postcode{28036}}

\author{Jose Manuel Gomez-Perez}
\email{jmgomez@expert.ai}
\orcid{0000-0002-5491-6431}
\affiliation{%
  \institution{expert.ai}
   \streetaddress{Poeta Joan Maragall 3}
  \city{Madrid}
  \country{Spain}
  \postcode{28036}}

\author{Jose Antonio Martínez-Heras}
\email{jose.martinez@solenix.ch}
\affiliation{%
  \institution{Solenix}
   \streetaddress{Spreestraße 3}
  \city{Darmstadt}
  \country{Germany}
  \postcode{64295}}
  
\author{Alessandro Donati}
\email{alessandro.donati@esa.int}
\affiliation{%
  \institution{ESA - ESOC}
  \city{Darmstadt}
  \country{Germany}
 }
  
\author{Ilaria Roma}
\email{Ilaria.Roma@esa.int}
\affiliation{%
  \institution{ESA - ESTEC}
  \city{Noordwijk}
  \country{Netherlands}
 }

\renewcommand{\shortauthors}{Garcia-Silva, et al.}

\begin{abstract}
We present SpaceQA, to the best of our knowledge the first open-domain QA system in Space mission design. SpaceQA is part of an initiative by the European Space Agency (ESA) to facilitate the access, sharing and reuse of information about Space mission design within the agency and with the public. We adopt a state-of-the-art architecture consisting of a dense retriever and a neural reader and opt for an approach based on transfer learning rather than fine-tuning due to the lack of domain-specific annotated data. Our evaluation on a test set produced by ESA is largely consistent with the results originally reported by the evaluated retrievers and confirms the need of fine tuning for reading comprehension. As of writing this paper, ESA is piloting SpaceQA internally. 
\end{abstract}



\maketitle

\section{Introduction}
\label{sec:intro}
The definition and assessment of future space missions or spacecraft concepts at the European Space Agency (ESA) is undertaken by a multidisciplinary group of experts at the Concurrent Design Facility (CDF), where concurrent engineering principles are applied to speed up the design process while ensuring consistent and high quality results \cite{Bandecchi1999ConcurrentEA}. This group of experts produces studies that establish the technical, programmatic, and economic feasibility of ESA's endeavours ahead of industrial development. Since its inception in 1998, the CDF has performed more than 250 studies and their associated reports. Typically, a CDF report\footnote{Public CDF reports available at: \url{https://www.esa.int/Enabling_Support/Space_Engineering_Technology/CDF/CDF_Reports}} is a long (between 200 to 300 pages) document in English that covers a large variety of technical topics related to the mission itself, the systems embedded in the mission, their configuration, payload, service module, ground segment and operations, and technical risk assessment, to name some of the most common ones. Finding specific pieces of information in such long documents using traditional search engines is a cumbersome, prone-to-error process. First, a keyword-based query is typically issued to retrieve candidate documents where the information being sought may be contained. Then, since the exact location of the answer to such query is not specified, such documents need to be manually inspected to find the specific answer.

\begin{figure}[t]
    \centering
    \includegraphics[width=0.49\textwidth]{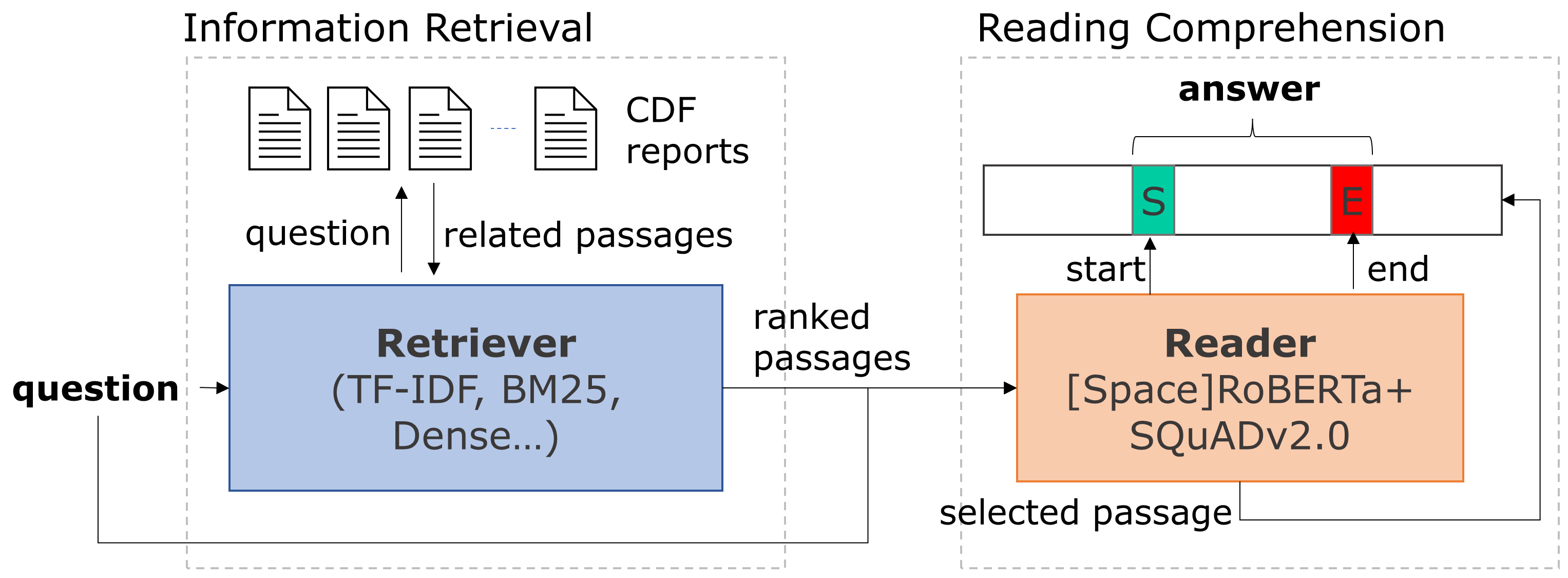}
    \caption{SpaceQA - Main components.}
   \label{fig:arch}
\end{figure}

To address these shortcomings, we formulate this problem as an open-domain question answering (QA) task~\cite{prager2006open}, which aims to answer a natural language question on a collection of large unstructured documents. As described in~\cite{chen-yih-2020-open}, open-domain QA has been a longstanding problem in natural language processing, information retrieval and related fields~\cite{Voorhees99thetrec-8,moldovan-etal-2000-structure,brill-etal-2002-analysis,Ferrucci_Brown_Chu-Carroll_Fan_Gondek_Kalyanpur_Lally_Murdock_Nyberg_Prager_Schlaefer_Welty_2010}. However, based on recent advances in neural reading comprehension, open-domain QA systems have experimented an accelerated evolution in the last few years. Complex pipelines involving many different components such as question processing, document/passage retrieval, and answer processing, have been replaced with modern approaches that combine traditional IR techniques and neural reading comprehension models~\cite{chen-etal-2017-reading,yang-etal-2019-end-end,DBLP:conf/emnlp/MinCHZ19}. 

In this paper, we present SpaceQA\footnote{Demo+video: \url{https://esatde.expertcustomers.ai/SpaceQA} user/pass: sigir/s1g1r2022!}, to the best of our knowledge the first implementation of an open-domain QA system for Space mission design\footnote{Source code and data: \url{https://github.com/expertailab/SpaceQA}}. As the first system of its kind in this domain, SpaceQA needed to face important challenges and limitations. Space documents and especially CDF reports are extremely technical documents that use a particularly domain-specific terminology~\cite{Berquand2020SpaceMD}. Plus, although English is an official language at ESA, very few of the authors of these documents are native speakers. However, the most limiting factor for the development of SpaceQA was the scarcity of domain-specific resources that we could use in combination with state-of-the-art architectures. In spite of promising recent work in transformer-based language models for Space science and engineering~\cite{Berquand2021SPaceRoBERTA}, the absolute lack of annotated data for open-domain QA in this domain prevents a strategy based on fine-tuning. Therefore, we decide to follow a transfer learning approach, leveraging pre-trained models that were fine-tuned on available datasets for similar tasks, and optimize them for the task at hand.  

The remainder of the paper is as follows. Section~\ref{sec:pre} describes how we preprocess and clean the PDF documents of the  CDF reports in our collection, extracting the text and preparing it for ingestion. Next, section~\ref{sec:app} describes the alternative approaches evaluated to implement the passage retrieval and reading comprehension components in the SpaceQA architecture (see figure~\ref{fig:arch}) and section~\ref{sec:eval} evaluates them on a manually crafted test set provided by ESA. We present an implementation of the SpaceQA system in section~\ref{sec:sys}, illustrating its behavior with several examples of Space mission design questions and their corresponding answers. In section~\ref{sec:rel}, we provide an overview of related work. 
Finally, section~\ref{sec:conc} concludes the paper and discusses impact and next steps at ESA.



\section{Document preprocessing}
\label{sec:pre}
To extract the text from the CDF reports we use Apache PDFBox.\footnote{Apache PDFBox https://pdfbox.apache.org} The output of PDFBox is a sequence of lines separated by new line characters that does not reflect the paragraph structure of the documents. To extract the paragraphs we remove all the new line characters that are not followed by another new line and split the whole text by the remaining new lines, obtaining the paragraphs that we use as passages. However, some of the passages extracted from the CDF reports are not relevant for SpaceQA, such as headers and footers, the cover page, or the items contained in the table of content. To filter such data we remove passages with less than $n$ characters. We empirically define $n=100$. While the number of CDF reports in our collection is not particularly large, the number of paragraphs in such documents is considerable. In total, we extracted 31,464 passages from a collection of 50 CDF reports. 

\section{Approach}
\label{sec:app}
The goal of SpaceQA is to find the answer to a factual question about Space mission design in a segment of text or span from a collection of CDF reports. Since answers need to be extracted from passages from a set of documents, we can allocate SpaceQA in the category of extractive~\cite{rajpurkar2018know} and open-domain~\cite{prager2006open} QA systems. 

As shown in figure~\ref{fig:arch}, SpaceQA follows a two-stage retriever-reader framework: i) a passage retriever component finding the passages that may contain an answer to the question from a collection of CDF reports and ii) a reader component finding the answer in some of such candidate passages. Keeping in mind the challenges mentioned in section \ref{sec:intro}, for the retriever we experiment with different methods including traditional sparse vector space methods based on TF-IDF and ranking methods including BM25 or cosine similarity, as well as dense representations using bi-encoders, while the reader is based on state-of-the-art reading comprehension models built on modern transformer architectures. 


\subsection{Passage Retrieval}
To retrieve the list of passages containing the answer to a question we consider several models. We start with sparse vectors using TF-IDF, and cosine similarity and BM25~\cite{robertson2009probabilistic} as ranking functions. We use Elasticsearch to index the passages extracted from the CDF reports, and set the search engine to use either cosine similarity or BM25 as ranking function. In addition, we use dense representations generated with transformer bi-encoders~\cite{reimers-gurevych-2019-sentence} to encode questions and passages and calculate their similarity  using vector dot product. Dense representations are able to capture semantic relations between words that are not possible by means of bag-of-words models used in the sparse representations. We evaluate the following alternative approaches to dense passage retrieval. 

\textbf{Dense Passage Retrieval (DPR)}~\cite{karpukhin2020dense} uses two BERT encoders and its loss function minimizes the distance between the CLS token representation of question and relevant passages. DPR was trained on  the Natural Questions corpus NQ  \cite{kwiatkowski-etal-2019-natural}. The NQ corpus consists of 307,373 queries issued to Google search engine and the answers are spans in Wikipedia articles. 

\textbf{ColBERT}~\cite{KhattabColBERT}  adapts BERT for efficient retrieval by adopting a late interaction architecture to compare bag of embeddings representing the question and the passage. The implementation\footnote{ \url{https://github.com/terrierteam/pyterrier_colbert}} that we use is fine tuned on the MS MARCO passage ranking dataset~\cite{bajaj2016msmarco}. This dataset contains 1 million questions extracted from Bing query logs and 8.8 million passages, extracted from 3.5 million Web pages,  which provide the necessary information to answer the questions.

Finally, \textbf{CoCondenser} \cite{gao-callan-2021-condenser}, a bi-encoder pre-training architecture
. This architecture adds a condenser head on top of the transformer that receives input from the output of an early layer for the tokens, while the input of the CLS is the output of the previous layer. Authors claim that this architecture enable the CLS representation to focus on the global meaning of the input text. We test a CoCondenser\footnote{\url{https://github.com/texttron/tevatron/tree/main/examples/coCondenser-marco}} trained on the MS Marco passage retrieval task. 

The three bi-encoders DPR, ColBERT and CoCondenser have reported state-of-the-art results in passage retrieval datasets. We use the bi-encoders to encode the passages extracted from the CDF reports and store the passage embeddings in FAISS~\cite{JohnsonFAISS}. FAISS is an  efficient library for similarity search and clustering of dense vectors. In inference time we use the bi-encoders to encode the question and then query the FAISS index for the most similar passages. 

\subsection{Reading comprehension}
Once a reduced set of the top k potential passages have been identified by the retriever, a neural reader attempts to spot the answer to the question as text spans in any of the passages. The neural reader assigns a score to each of the extracted candidate spans and the span with the highest score is returned as the answer. Due to the lack of a question answering dataset for the space domain to train the neural reader, we resort to the widely used Stanford Question Answering dataset (SQuAD2.0) \cite{rajpurkar2018know}. SQuAD2.0 contains questions, answers, and passages extracted from Wikipedia. In total there are 150,000 questions, 50,000 of which are not answerable in the given passage.

The first neural reader we use is a \textbf{RoBERTa}~\cite{liu2019roberta} model fine-tuned on  SQuAD2.0. Since RoBERTa was pre-trained on a general purpose corpus, there could be a vocabulary mismatch between the set of questions in the Space mission design domain in our evaluation dataset and the language model that could affect its performance. Thus, we also evaluate \textbf{SpaceRoBERTa} \cite{Berquand2021SPaceRoBERTA}, a version of RoBERTa pre-trained on documents from the Space Domain. This model started from a RoBERTa base model and was further trained on a 14.3 GB corpus of publications abstracts, books, and Wikipedia pages related to space systems.

We evaluate existing versions of both RoBERTa base\footnote{ \url{https://huggingface.co/deepset/roberta-base-squad2}} and large\footnote{\url{https://huggingface.co/phiyodr/roberta-large-finetuned-squad2}} available in Hugging Face, which had already been fine tuned on SQUAD2.0. However, in the case of SpaceRoBERTa,\footnote{ \url{https://huggingface.co/icelab/spaceroberta}} we carry out the fine-tuning process. To fine-tune SpaceRoBERTa in SQuAD2.0 we use the span prediction task originally proposed in BERT~\cite{devlin-etal-2019-bert}. That is, the question and passage are represented as a single sequence separated by the [SEP] token, and the model is trained to the detect the start and end positions of the answer span. To deal with not answerable questions, we represent the answer span with start and end at the [CLS] token. We fine-tune the model for 2 epochs with a learning rate of 3e-5, and a batch size of 10\footnote{Due to hardware limitations, this version of SpaceRoBERta fine tuned on SQuAD2.0 was trained with a smaller batch size (10 vs. 96 for the other two models).}

\section{Evaluation}
\label{sec:eval}
We use a dataset of factual questions produced by ESA, with answers and paragraphs extracted from CDF reports where the answer to the question appears. The set contains 60 questions, answers, and corresponding paragraphs. While this dataset is useful to evaluate the reading comprehension model by testing whether the right answer is extracted for a question, it is limited to evaluate the passage retrieval module since potentially more than one paragraph can contain the answer for a given question. Thus, we manually search for additional paragraphs containing the answer to the question and extend the dataset with up to 5 more paragraphs for each question. 

\begin{table}[ht!]
  \caption{Evaluation of the retrievers.}
  \label{tab:retrievers}
  \begin{tabular}{ccccl}
    \toprule
    Retriever & R@10 & MRR@10 & Accuracy\\
    \midrule
    TF-IDF & 0.252 & 0.395 & 0.483  \\ 
    BM25 & 0.326 & 0.254 & 0.55  \\ \hline
    DPR & 0.218  & 0.170 & 0.35 \\
    ColBERT & \textbf{0.4898}  & \textbf{0.560} & \textbf{0.717} \\
    CoCondenser & 0.354 & 0.404 & 0.583  \\
  \bottomrule
\end{tabular}
\end{table}

\begin{table}[ht!]
  \caption{Evaluation of the readers.}
  \label{tab:readers}
  \begin{tabular}{cccc}
    \toprule
    Model & Precision & Recall & F-Score \\
    \midrule
    RoBERTa base & 0.774 & \textbf{0,751} & \textbf{0.762} \\
    RoBERTa large & 0.629  & 0.664 & 0.646 \\ 
    SpaceRoBERTa & \textbf{0.816} & 0.671 & 0.737 \\
  \bottomrule
\end{tabular}
\end{table}

Since the area of open-domain QA in Space mission design is relatively unexplored, and hence there is a lack of systems to compare against, we focus our evaluation in understanding the current limitations of SpaceQA. To this purpose, rather than overall end-to-end performance we are particularly interested in each of the steps involved in SpaceQA individually
. Table~\ref{tab:retrievers} shows the evaluation results of the retrievers using recall and mean reciprocal rank at 10. We also measure the accuracy of the retrievers to find within the 10 top positions at least one passage containing the answer. The results obtained in our test set seem to be consistent with those reported in the original papers (except DPR), supporting in this case our decision to adopt a transfer learning approach. In all three metrics, ColBERT was the best retriever, followed by CoCondenser. Training on MS MARCO seems to be key when reusing such retrievers. Trained on NQ, DPR performs last in our test set. 

Regarding the readers, it has been reported that for natural language questions there are often a number of acceptable answers as well as a genuine ambiguity in whether or not an answer is acceptable~\cite{kwiatkowski-etal-2019-natural}. For example, for the question \textit{why is the rover top part larger than the bottom part?} an acceptable answer is \textit{to support the solar panels} but \textit{to support the solar panels stack namely 910mm by 500mm} is acceptable, too. A span exact match evaluation could result on low and discouraging results, which do not reflect the actual performance of the reader. Thus, we opt for a token based evaluation of the reader where we compare the tokens in the span extracted by the reader against the lists of tokens in the ground truth answer. Evaluation results are presented in table \ref{tab:readers}. Unlike in the case of the retrievers, these results illustrate the impact of the lack of domain-specific data for fine-tuning and suggest considerable room for improvement. As a reference, the current SotA in extractive QA over SQuAD2.0,\footnote{\url{https://rajpurkar.github.io/SQuAD-explorer}} is currently above 93.2 F1. Interestingly, RoBERTa base was the best reader overall, followed by  SpaceRoBERTa. SpaceRoBERTa precision is higher than RoBERTa base, however its recall is lower.

\begin{figure*}[ht]
    \centering
    \includegraphics[width=\textwidth]{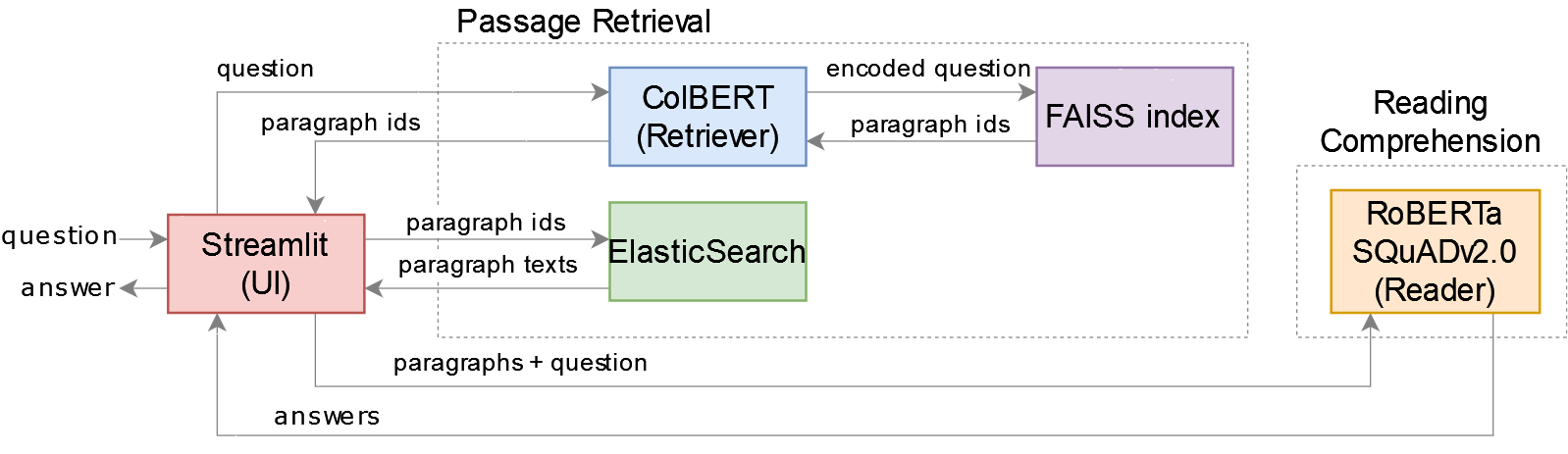}
    \caption{SpaceQA - System architecture.}
   \label{fig:System-implementation}
\end{figure*}

\section{System description}
\label{sec:sys}
We build an open-domain QA system using the best performer retriever and reader from the evaluation in section~\ref{sec:eval}: ColBERT as retriever and RoBERTa-base fine-tuned on SQuAD2.0 as reader. Figure \ref{fig:System-implementation} shows a high-level view of system architecture and how the different component interacts to extract answers for user questions. In a batch process we index in Elasticsearch the text in the passages extracted from the CDF reports, and the passage encodings generated with ColBERT in FAISS. 

At runtime when an user asks a question in the frontend  the system uses ColBERT to encode the question and retrieve from FAISS the 10 most relevant passage ids. Then the passage ids are used to retrieve from ElasticSearch the text of the passages. Next, the pairs of question and relevant passages are processed by the reader and the extracted answers along with the score are returned. However, only answers with a score above an empirical defined threshold of 0.5 are shown directly to the user. To see answers with a lower score the user is prompted first with an warning message about the low certainty of the answers to be displayed. In the infrastructure side we deploy the system in a machine with 32BG RAM, 1TB SSD, intel i7 CPU, and NVIDIA GeForce 1080Ti GPU. Both the reader and FAISS run on the GPU while the rest of the components use the CPU. 

\begin{figure}[htbp]
  \centering
  \includegraphics[width=\columnwidth]{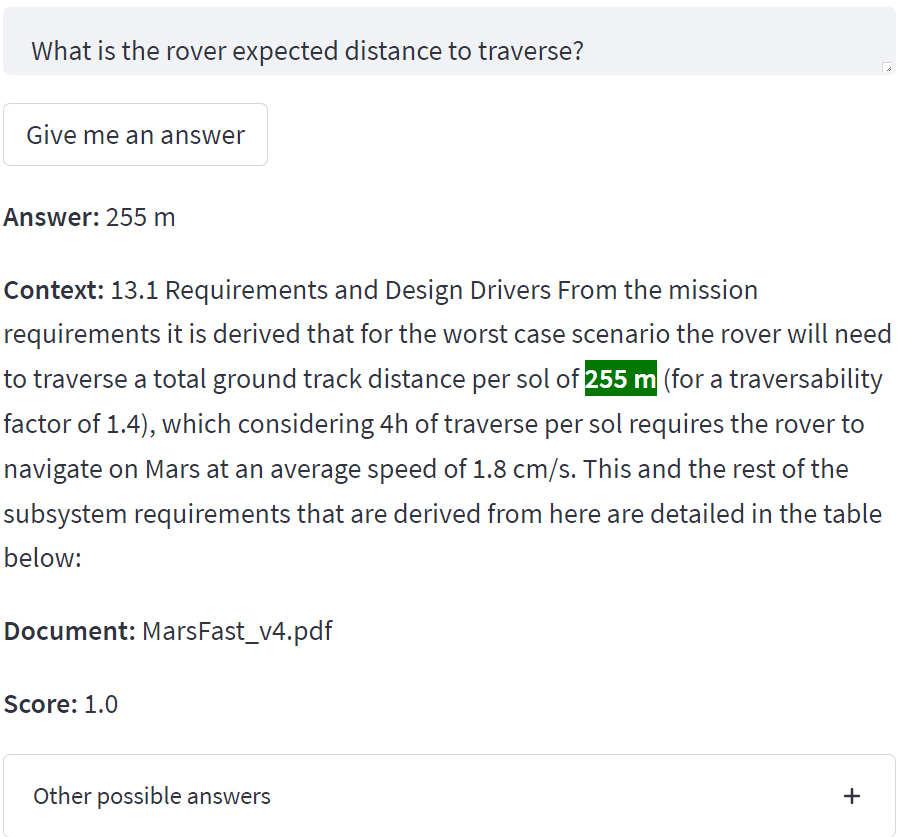}
  \caption{Screenshot of the SpaceQA user interface.}
  \label{fig:screenshot}
\end{figure}

Figure \ref{fig:screenshot} shows a screenshot of the SpaceQA web application. When the user poses a question, the system displays the answer with the highest score and the passage from where it was extracted. We highlight the answer in the passage to easily spot the answer context. We also display and link the source document of the passage and the answer score, obtained by multiplying the probabilities of the span start and end tokens generated by the reader. In addition, if the system extracts other answers these are also displayed sorted by score under the option "other possible answers". When the score of an answer is below 0.5, we show a warning message reporting that the system confidence in the answer is low and the user is required to actively click on the option to see it. If the reader does not find any answer, we also inform the user.

Since most of the users are used to keyword-based queries in traditional search engines we provide a list of predefined questions so that they can experiment with the system features and see examples of the type of questions the system handles. We also provide the option to focus the question on a specific CDF report. This is particular important for questions that lack enough details and return valid answers from different reports. For example, asking a question about a vehicle or instrument that can be part of different missions can bring potential irrelevant results.

\begin{table}[htbp]
  \centering
  \caption{Questions and answers extracted by SpaceQA}
   \resizebox{\columnwidth}{!}{ 
    \begin{tabular}{l}
    1. Which launcher will athena use? \\
     \hspace{3mm} Ariane 5 \\
    2. What wavelengths can be observed by NG-CryoIRTel? \\
    \hspace{3mm} 20-200µm \\
    3. What is the purpose of the tunable laser spectrometer?  \\
    \hspace{3mm} detect trace concentration of water and volatiles \\
    4. How is the ATHENA mirror structure manufactured?  \\
    \hspace{3mm} 3D-printing \\
    5. Where will NG-CryoIRTel be launched from?  \\
    \hspace{3mm} Tanegashima Space Centre \\
    6. Where is the panoramic camera mounted?  \\
    \hspace{3mm} a deployable mast \\
    7. When can dust storms occur during the MarsFAST mission? \\
    \hspace{3mm} at any time \\
    8. How long will the NG-CryoIRTel mission last?  \\
    \hspace{3mm} at least 5 years \\
    \end{tabular}%
    }
  \label{tab:examples}%
\end{table}%

In table \ref{tab:examples} we show some example questions and the answers produced by SpaceQA. The system deals effectively with different types of questions (e.g., what, which, where,  when, and how), and provides appropriate answers in the form of rockets, units of measure, descriptions, locations, things, and time periods. Nevertheless, there are questions for which the system does not provide an answer, or the answer is wrong. While some of these are poorly specified questions\footnote{For example, the question \textit{Why can the sample material not be exposed to daylight?} does not add any information about sample material.} lacking details that can help to increase the confidence of the reader, others are just not properly answered. Recall that in our system we use models that were pre-trained  on general purpose corpora and fine-tuned on MS MARCO and SQUAD2.0, none of them rich in the Space domain. Therefore, a natural next step to improve the SpaceQA system is to generate a QA dataset for the space domain and then use this dataset to fine-tuned the reader. 

\section{Related Work}
\label{sec:rel}
Open-domain QA has evolved rapidly in the last few years based on recent advances in the combination of IR and reading comprehension~\cite{chen-etal-2017-reading,yang-etal-2019-end-end,DBLP:conf/emnlp/MinCHZ19} and even end-to-end approaches~\cite{lee-etal-2019-latent,seo-etal-2019-real,Guu2020REALMRL,roberts-etal-2020-much}. The application of open-domain QA in Space mission design is new. However, due to the need of exploring large collections of documents, domains like Healthcare have been a traditional field of application~\cite{10.1145/3490238,Cao2011AskHermes,Cairns2011TheMC}. Recently, the conversation around COVID-19 has increased the need for modern systems that allow asking questions and receiving credible, scientific answers~\cite{levy-etal-2021-open}. Other systems~\cite{Zhou2021QA4Chem} rely on knowledge graphs to answer questions in domains with a large amount of structured data like Chemistry. 

\section{Conclusions}
\label{sec:conc}

This paper presents SpaceQA, the first open-domain QA system in Space mission design. SpaceQA has been developed under the auspices of ESA with the participation of the European Space Operations Center (ESOC) and the Research and Technology Center (ESTEC) as part of an initiative to facilitate the access, sharing and reuse of information about Space mission design within the agency and with the public. During the next months, ESA plans to pilot SpaceQA beyond the scope of the team that participated in the development of the system and collect feedback for further refinement. Such activities will include a data annotation campaign to gather an annotated dataset for reading comprehension in Space mission design,  identified in our evaluation as a clear future improvement. 

\begin{acks}
We thank to Patrick Fleith, Patrick Lux, Massimo Magazzu, and Stefano Scaglioni for their collaboration to develop SpaceQA. This work is funded by ESA under contract AO/1-10291/20/D/AH - “Text and Data Mining to Support Design, Testing and Operations”.
\end{acks}

\bibliographystyle{ACM-Reference-Format}
\bibliography{spaceqasigir}

\end{document}